\documentclass{ifacconf}

\usepackage{graphicx}      
\usepackage{natbib}        
\usepackage{multicol}
\usepackage{booktabs} 
\usepackage{multirow} 
\usepackage[T1]{fontenc}
\usepackage{siunitx} 
\usepackage{xcolor}  
\usepackage{soul}    

\usepackage[most]{tcolorbox} 
\usepackage{listings}       
\usepackage{enumitem}       

\definecolor{codegray}{rgb}{0.95,0.95,0.95}
\definecolor{codegreen}{rgb}{0,0.6,0}
\definecolor{codepurple}{rgb}{0.58,0,0.82}
\definecolor{codeblue}{rgb}{0,0,0.8} 

\lstdefinelanguage{json}{
    keywords={true, false, null},
    keywordstyle=\color{codeblue},
    string=[s]{"}{"},
    stringstyle=\color{codepurple},
    comment=[l]{//},
    commentstyle=\color{codegreen},
    morecomment=[s]{/*}{*/},
    numberstyle=\small,
    basicstyle=\ttfamily\small,
    showstringspaces=false 
}

\lstdefinestyle{json_style}{
    language=json, 
    backgroundcolor=\color{codegray},
    breakatwhitespace=false,
    breaklines=true,
    captionpos=b,
    keepspaces=true,
    showspaces=false,
    showtabs=false,
    tabsize=2,
    frame=single,
    framerule=0pt,
    rulecolor=\color{black},
    numbers=left,
    numberstyle=\tiny\color{gray}
}

\begin{document}
\begin{frontmatter}

\title{Scale, Don't Fine-tune: Guiding Multimodal LLMs for Efficient Visual Place Recognition at Test-Time \thanksref{footnoteinfo}} 

\thanks[footnoteinfo]{Correspond Author: Jin Wu (e-mail: wujin@ustb.edu.cn)}

\author[First]{Jintao Cheng} 
\author[Second]{Weibin Li} 
\author[Second]{Jiehao Luo}
\author[Second]{Xiaoyu Tang}
\author[Third]{Zhijian He}
\author[Fourth]{Jin Wu}
\author[Fourth]{Yao Zou}
\author[First]{Wei Zhang}

\address[First]{Hong Kong University of Science and Technology, Hong Kong, China 
(e-mail: jchengau@connect.ust.hk, eeweiz@ust.hk)}
\address[Second]{South China Normal University, Shanwei, Guangdong, China 
(e-mail: 20228131086@m.scnu.edu.cn, 20228132034@m.scnu.edu.cn)}
\address[Third]{Shenzhen Technology University, Shenzhen, Guangdong, China
   (e-mail: hezhijian@sztu.edu.cn)}
\address[Fourth]{University of Science and Technology Beijing, 
   Beijing, China 
   (e-mail: wujin@ustb.edu.cn, zouyao@ustb.edu.cn)}


\begin{abstract}
Visual Place Recognition (VPR) has evolved from handcrafted descriptors to deep learning approaches, yet significant challenges remain. Current approaches, including Vision Foundation Models (VFMs) and Multimodal Large Language Models (MLLMs), enhance semantic understanding but suffer from high computational overhead and limited cross-domain transferability when fine-tuned. To address these limitations, we propose a novel zero-shot framework employing Test-Time Scaling (TTS) that leverages MLLMs' vision-language alignment capabilities through Guidance-based methods for direct similarity scoring. Our approach eliminates two-stage processing by employing structured prompts that generate length-controllable JSON outputs. The TTS framework with Uncertainty-Aware Self-Consistency (UASC) enables real-time adaptation without additional training costs, achieving superior generalization across diverse environments. Experimental results demonstrate significant improvements in cross-domain VPR performance with up to 210$\times$ computational efficiency gains.
\end{abstract}


\end{frontmatter}

\section{Introduction}

Visual Place Recognition (VPR) represents a cornerstone capability in computer vision and robotics, enabling systems to determine whether a location has been previously visited based solely on visual input. This fundamental capability underpins critical applications ranging from loop closure detection in Simultaneous Localization and Mapping (SLAM) \cite{MF-MOS} to autonomous navigation and long-term robot deployment in dynamic environments.

The evolution of VPR has progressed from traditional handcrafted descriptors like SIFT by~\cite{14cruz2012scale} to sophisticated deep learning approaches, accroding to~\cite{16lowry2015visual, 17zhang2021visual}. While early methods offered computational efficiency, they struggled with environmental variations and appearance changes due to their fixed feature representations. The advent of deep learning fundamentally transformed the field by introducing learnable representations capable of adapting to complex visual conditions. CNNs demonstrated superior performance through hierarchical feature learning, with landmark methods such as Patch-NetVLAD by~\cite{57hausler2021patch}, MixVPR by~\cite{58ali2023mixvpr}, and EigenPlaces by~\cite{59berton2023eigenplaces} successfully capturing both geometric and semantic information. These approaches significantly outperformed traditional methods by learning robust feature representations directly from data. Building upon this, Vision Foundation Models (VFMs) have recently provided unprecedented VPR capabilities through large-scale pre-training on massive datasets.

\begin{figure}[ht]
  \centering
  \includegraphics[width=0.49\textwidth]{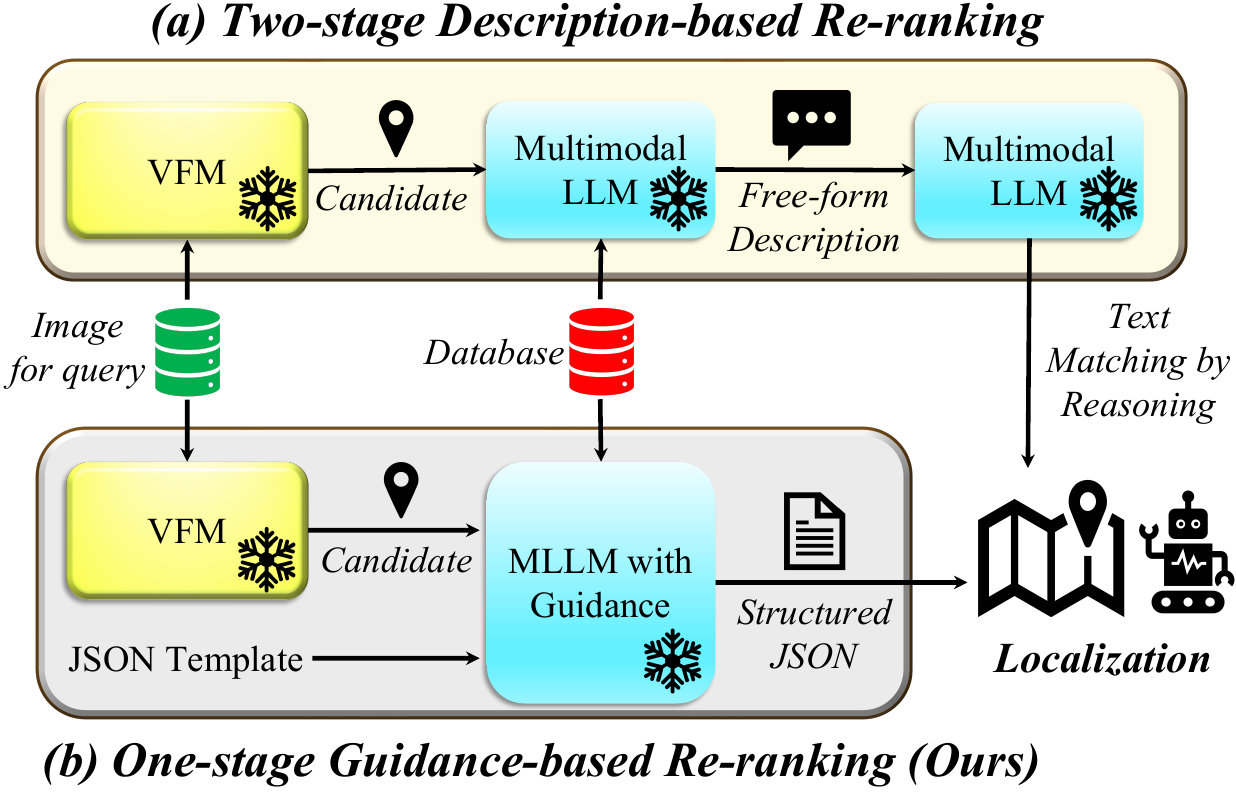}
  \caption{\textbf{Comparison between traditional Description-based method and our proposed approach for VPR}. Our method better leverages MLLM's vision-language alignment capabilities through one-stage direct multimodal reasoning, while structured JSON output enables more controlled thinking pathways and reduces computational overhead compared to free-form text generation.}
  \label{fig:abstract}
\end{figure}

\begin{figure*}[ht]
  \centering
  \includegraphics[width=1\textwidth]{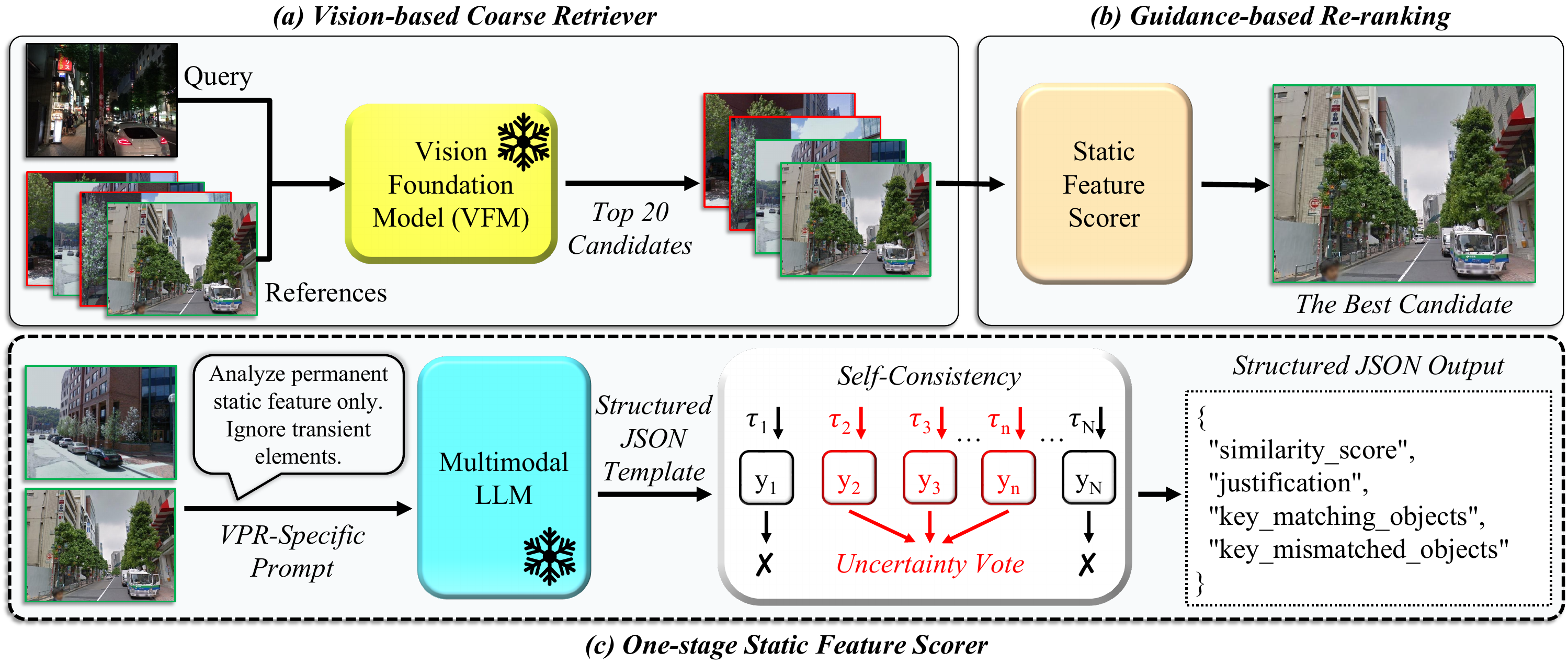}
  \caption{\textbf{Overall architecture of our proposed framework.} The framework integrates vision-based coarse retrieval with Guidance-based re-ranking, where MLLMs generate structured JSON outputs through VPR-specific prompts and self-consistency mechanisms.}
  \label{fig:archi}
\end{figure*}

A recent line of work incorporates Vision-Language Models (VLMs) into place recognition to transcend the limitations of pure visual feature matching, like ProGeo by \cite{hu2024progeo}, GeoClip by \cite{vivanco2023geoclip} and AddressClip by \cite{xu2024addressclip}. By leveraging the capacity of VLMs for high-level semantic understanding and generalization, these approaches aim to achieve more robust performance amidst severe appearance variations in challenging environments. These studies initiated the transition of VPR from a purely visual task to a multimodal one, utilizing information from other modalities to aid in the retrieval process. However, the aforementioned methods all require fine-tuning, which is in fundamental conflict with the core requirement of place recoginition: to operate reliably across diverse and previously unseen locations.

\par

Taking a step further, NAVIG by \cite{zhang2025navig} has begun to employ Multimodal Large Models (MLLMs) for the place recognition task, though this method still relies on fine-tuning. Subsequently, LLM4VPR by \cite{lyu2024tell} have successfully leveraged MLLMs in a zero-shot fashion by reformulating the VPR task from image retrieval to text generation, achieving greate performance. This approach follows a coarse-to-fine architecture. It begins by using VFM to perform an initial coarse ranking and retrieve the top-N candidates. The fine-grained re-ranking is then handled by MLLM, which first translates each candidate image into a detailed textual description. These descriptions are subsequently passed to LLM that performs the final re-ranking based solely on the textual information. Given that the core logic hinges on this intermediate textual representation, we term this the Description-based approach.
\par
However, by converting images into text, the re-ranking process loses access to the rich visual data, making its success entirely dependent on the quality of the initial textual generation. This dependency is a critical point of failure, especially for ambiguous scenes where MLLMs may struggle to articulate the decisive visual cues necessary for accurate re-ranking. Moreover, the generative step is computationally expensive.
\par
To address these limitations, we introduce Guidance-based method, which bypasses the explicit vision-to-text conversion entirely. It is built on the principles of two Test-Time Scaling (TTS) strategies: Chain-of-Thought (CoT) and Self-Consistency. We use a meticulously crafted CoT prompt to guide the MLLM's reasoning process directly on the visual input, preserving information fidelity and drastically reducing computational overhead. We further enhance this framework with our proposed Uncertainty-Aware Self-Consistency (UASC), a mechanism that gauges score reliability via heteroscedastic uncertainty, empowering the model to make more robust decisions when faced with challenging candidates that yield similar scores. As shown in Fig.~\ref{fig:abstract}, our Guidance-based framework contrasts with previous Description-based methods by utilizing structured prompts for direct similarity scoring, achieving superior performance through simplified single-stage re-ranking processing. The key contributions of this paper are summarized as follows:
\begin{enumerate}
    \item The first work to apply TTS strategies to zero-shot MLLMs for VPR. By integrating TTS within coarse-to-fine framework, our method demonstrates exceptional generalization and off-the-shelf usability without requiring any fine-tuning.
    \item A novel Guidance-based framework is proposed that leverages MLLM's vision-language alignment capabilities through structured prompts, enabling direct end-to-end similarity scoring without intermediate text generation;
    \item Uncertainty-Aware Self-Consistency is developed that dynamically calibrates predictions by quantifying aleatoric uncertainty during inference;
    \item Competitive performance is achieved with significant computational efficiency gains, demonstrating substantial improvements in both accuracy and processing speed compared to existing approaches.
\end{enumerate}

\section{Related Work}

\subsection{Visual Place Recognition}
Visual Place Recognition, together with complementary approaches such as the LiDAR-based Place Recognition method by \citet{overlapmamba}, represents a core methodology in place recognition and has evolved from traditional handcrafted descriptors to sophisticated deep learning architectures. Early methods relied on engineered features like SURF by~\cite{15bay2008speeded}, which offered computational efficiency but struggled with environmental variations and semantic understanding. The foundational survey by~\cite{16lowry2015visual} established the theoretical framework for VPR, while~\cite{17zhang2021visual} later provided a comprehensive review from the deep learning perspective. The introduction of CNN-based approaches marked a paradigmatic shift, with pooling-based methods like NetVLAD becoming cornerstone techniques for global descriptor generation. Recent advances have embraced Transformer architectures, leveraging self-attention mechanisms to capture long-range spatial dependencies. \cite{65xu2023transvlad} developed TransVLAD using multi-scale attention-based global descriptors, while \cite{67ali2024boq} introduced BoQ, reformulating place recognition as a learnable query-based problem. \cite{68lu2024cricavpr} presented CricaVPR with cross-image correlation-aware representation learning, demonstrating superior performance across diverse datasets. Despite these advances, current VPR methods face challenges in computational efficiency, domain adaptation, and handling extreme environmental variations, particularly when deployed in resource-constrained scenarios or novel environments significantly different from training conditions.

\subsection{Test-Time Scaling Approaches for LLM}

Test-Time Scaling has emerged as a promising paradigm for enhancing model performance through dynamic computation allocation during inference, inspired by human cognitive processes that engage deeper thinking for complex problems. Early approaches focused on inference-time methods, with Chain-of-Thought (CoT) prompting by~\cite{wei2022chain} demonstrating that explicit reasoning steps significantly improve performance on complex tasks. Self-Consistency mechanisms by~\cite{wang2023h} and Best-of-N sampling by~\cite{brownqiu2024treebon} further enhanced reliability through multiple candidate generation and selection strategies. Tree-of-Thought (ToT) frameworks by~\cite{yao2023tree} advanced this direction by modeling reasoning as systematic tree search processes. Recent breakthroughs, such as DeepSeek's R1 by~\cite{marjanovic2025deepseek}, represent hybrid scaling strategies that integrate training-time reasoning development with test-time dynamic adjustment capabilities. Advanced search methods including MCTS by~\cite{feng2023alphazero} and adaptive computational allocation strategies by~\cite{snell2024scaling} dynamically adjust scaling based on problem complexity. The key advantage of TTS for VPR lies in addressing the fundamental performance-generalization trade-off. Unlike fine-tuning approaches that create domain-specific models with limited transferability, TTS preserves the generalization capabilities of pre-trained models while achieving superior performance through intelligent inference-time optimization. This enables immediate deployment in novel environments without costly adaptation procedures, making it particularly attractive for robotics applications.

\section{Methodology}
\subsection{Overall Framework}
Our proposed framework introduces a novel approach for MLLM-based VPR re-ranking, as is shown in Fig. 2. The methodology is centered on efficiently leveraging the advanced multimodal reasoning and instruction-following capabilities of MLLMs. It consists of two primary components: (1) a single-stage, Guidance-based method for pairwise similarity scoring, and (2) an Uncertainty-Aware Self-Consistency (UASC) strategy to calibrate and enhance the robustness of these scores.

\subsection{Guidance-Based method}
The foundational insight of our approach is to reframe the re-ranking task not as a visual-to-text conversion problem, but as an instruction-guided, multimodal reasoning task. Instead of requiring the MLLM to first generate exhaustive textual descriptions of images and then reason over that text in a second stage, our method guides the model to perform a direct, end-to-end comparison in a single pass. This is achieved through meticulously engineered, VPR-CoT prompting.
\par
For each query-candidate image pair $(I_q, I_{c,k})$, we construct a VPR-CoT multimodal prompt $\Pi_k$, which guides the model through three key components:
  \begin{enumerate}
    \item \textbf{VPR Task Constraints:} The prompt explicitly instructs the model to analyze only permanent, static features and to systematically ignore all transient elements. This core constraint precisely aligns the model's analysis with the essential requirement of VPR: identifying the place itself, not its temporary contents.
    \item \textbf{Structured Reasoning \& Scoring Rubric:} We provide the model with a detailed scoring rubric that maps qualitative levels of visual similarity to a continuous numerical scale from 0.0 to 1.0. This not only transforms an abstract comparison task into a constrained quantitative evaluation but also guides the model's internal reasoning process.
    \item \textbf{Mandatory JSON output:} To ensure the output is length-controllable and interpretable, the prompt strictly mandates that the final response be a single, raw JSON object. This object must adhere to a predefined schema, containing the numerical \texttt{similarity\_score}, a textual \texttt{justification}, and lists of key matching and mismatched objects.
  \end{enumerate}

\par
By integrating these guiding instructions, the MLLM processes visual and textual information concurrently in a unified process to directly generate the final score for re-ranking. 

\subsection{Uncertainty-Aware Self-Consistency}
In Bayesian modeling, a model's predictive uncertainty can be divided into two primary types: epistemic uncertainty and aleatoric uncertainty. Our methodology focuses on the latter—aleatoric uncertainty—which captures the inherent noise and ambiguity present in the data itself.
\par
Aleatoric uncertainty can be further categorized as either homoscedastic, which remains constant for all inputs of a given task, or heteroscedastic, which varies with the input data. In our VPR task, the difficulty of matching different image pairs varies—some pairs are clear and unambiguous, while others are inherently ambiguous. Therefore, the inconsistency exhibited by the model during prediction is a form of data-dependent, or heteroscedastic, aleatoric uncertainty.

\par
To empirically estimate the aleatoric uncertainty for a given image pair $(I_q, I_{c,k})$, we utilize self-consistency strategy. We conduct $N$ independent stochastic sampling passes for each input prompt $\Pi_k$. For the $i$-th pass, the model $M$ generates a raw textual response $R_{k,i}$:
\begin{equation}
     R_{k,i} \sim M(\Pi_k; \Theta=\{\tau\}) 
\end{equation}
    This process yields a set of raw responses $S_{R,k} = \{R_{k,1}, R_{k,2}, \dots, R_{k,N}\}$.
    \par
We parse and validate each response in the raw set $S_{R,k}$. The \texttt{similarity\_score} from each response that contains a valid, numerical score is extracted. These scores form the final set of scores, denoted as $S_{\text{scores},k}$. We then compute the key statistical moments from this set.
  \begin{itemize}
    \item The mean $\mu_{s,k}$, which serves as the model's central estimate of the similarity score.
    \begin{equation}
             \mu_{s,k} = \frac{1}{|S_{\text{scores},k}|} \sum_{s \in S_{\text{scores},k}} s 
    \end{equation}

    \item The sample standard deviation $\sigma_{s,k}$, which serves as a direct quantification of the aleatoric uncertainty for this specific input.
    \begin{equation}
            \sigma _{s,k}=\left( \frac{1}{|S_{\mathrm{scores},k}|-1}\sum_{s\in S_{\mathrm{scores},k}}{(}s-\mu _{s,k})^2 \right) ^{\frac{1}{2}}
    \end{equation}
  \end{itemize}
We calibrate the central estimate by a penalty term proportional to the quantified aleatoric uncertainty. The calibrated score $s_{\text{calibrated}, k}$ is computed as:
\begin{equation}
     s_{\text{calibrated}, k} = \mu_{s,k} - \lambda \sigma_{s,k} 
\end{equation}
    where $\lambda$ is a hyperparameter that controls the strength of the uncertainty penalty. Finally, the score is clamped to the valid interval of [0, 1]:
    \begin{equation}
     s_{\text{final}, k} = \max(0, \min(1, s_{\text{calibrated}, k})) 
    \end{equation}
    This methodology ensures that only predictions with both a high mean score and high consistency (low aleatoric uncertainty) are prioritized in the final re-ranking.

\begin{figure}[t]
  \centering
  \includegraphics[width=0.48\textwidth]{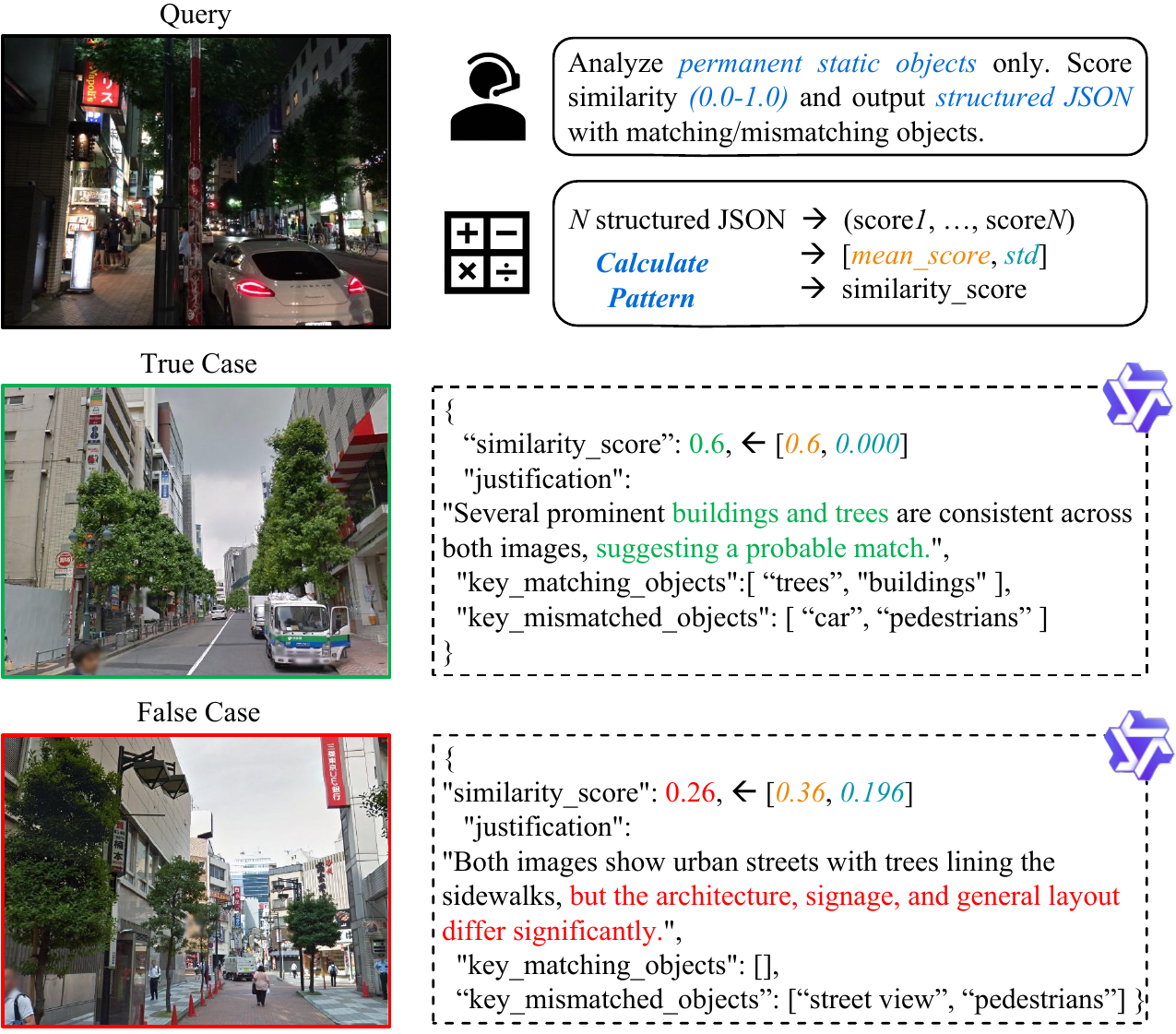}
  \caption{\textbf{Case Study of Guidance-based Similarity Scoring.} Examples showing our prompt, calculating pattern and structured JSON outputs with similarity scores and justifications. True matches achieve high scores of 0.6 while false matches receive low scores of 0.26 based on permanent static feature analysis.}
  \label{fig:case}
\end{figure}
\section{Experiment}
\subsection{Experiment Setup}
\subsubsection{Datasets}
We evaluated our approach on the Tokyo247 dataset by~\cite{Torii_2015_CVPR} and Pittsburgh30K dataset by~\cite{Torii_2013_CVPR}. Tokyo247 comprises street-view imagery captured across Tokyo under both daytime and nighttime conditions, and consists of 315 query images and approximately 76,000 reference images.The Pittsburgh30K dataset is a large-scale urban visual localization benchmark comprising approximately 30,000 street-view images captured across Pittsburgh, featuring diverse viewpoints and lighting conditions.

\subsubsection{Evaluation Metric}
We employ Recall@K (R@K) as our evaluation metric. For each query, a retrieval is deemed successful if at least one of the top-K retrieved candidates has a geographical location within m meters of the query's ground truth. The R@K score is then computed as the proportion of successfully retrieved queries to the total number of queries.

\subsubsection{Implementation Details}
We benchmark our proposed Guidance-based method against the training-free approach of traditional VFM and the Description-based method of VPR-LLM. The VPR-LLM baseline first retrieves the top-20 candidates per query using DINOv2 GeM, followed by re-ranking with a multimodal LLM. For the re-ranking module, we evaluate two models: Qwen2.5-VL-7B-Instruct (Qwen-7B) and Qwen2.5-VL-32B-Instruct (Qwen-32B). On the Pitts30k dataset, we exclusively test the Guidance-based Qwen-7B due to computational resource constraints. The prompt and implementation procedure for the Description-based method strictly follow the methodology outlined in LLM4VPR. All experiments are conducted without fine-tuning.

\subsection{Overall Performance}

\begin{table}[htbp]
\centering
\caption{Performance comparison on Tokyo247}
\label{tokyo}
\renewcommand{\arraystretch}{1.2}
\begin{tabular}{@{} l l S[table-format=1.4] S[table-format=1.4] S[table-format=1.4] @{}} 
\toprule
\textbf{Method}  & \textbf{Model}        & {\textbf{R@1}} & {\textbf{R@5}} & {\textbf{R@10}} \\
\midrule
\multirow{2}{*}{VFM} & DINOv2 CLS          & {60.00}          & {77.78}          & {82.54}           \\
                 & DINOv2 GeM          & {77.14}          & {91.43}          & {93.65}           \\
\midrule
\multirow{8}{*}{LLM-VPR} & \multicolumn{4}{l}{\textit{Description-Based}} \\
\cmidrule(l{2pt}r{2pt}){2-5} 
                 & \quad Qwen-7B  & {71.75}          & {86.03}          & {89.52}           \\
                 & \quad Qwen-32B & {74.60}          & {92.06}          & {94.60}           \\[0.5ex]
\cmidrule(l{2pt}r{2pt}){2-5}
                 & \multicolumn{4}{l}{\textit{Guidance-Based (ours)}} \\
                 
\cmidrule(l{2pt}r{2pt}){2-5} 
                 & \quad Qwen-7B       & {83.81}         & {91.75}         & {93.65}          \\
                 & \quad Qwen-7B-UASC  & {84.44}         & {91.75}         & {93.33}          \\
                 & \quad Qwen-32B      & {89.52}         & {\textbf{94.60}}         & {\textbf{94.62}}          \\
                 & \quad Qwen-32B-UASC & {\textbf{91.11}}         & {\textbf{94.60}}         & {\textbf{94.62}}          \\
\bottomrule
\end{tabular}
\end{table}

As shown in Table \ref{tokyo}, our results on the Tokyo247 dataset indicate that GeM pooling serves as a significantly more effective global descriptor than the [CLS] token. Under the Description-based paradigm, both Qwen-7B and Qwen-32B underperform the DINOv2 GeM baseline. This finding deviates considerably from the results reported in LLM4VPR by~\cite{lyu2024tell}, a discrepancy we attribute to their use of the more powerful GPT-4V model. This suggests that the Description-based approach is highly demanding of the underlying model's capabilities and may not fully unleash its potential.
\par
In contrast, our proposed Guidance-based method yields substantial performance improvements. Compared to the DINOv2 GeM baseline, Qwen-7B and Qwen-32B achieve R@1 gains of 6.67 and 12.42 percentage points, respectively. Furthermore, by incorporating the UASC strategy, the Qwen-32B model achieves a remarkable R@1 of 91.11\%, highlighting the significant potential of our approach.

\begin{table}[t]
\centering
\caption{Performance comparison on Pitts30k}
\label{tab:pitts30k}
\renewcommand{\arraystretch}{1.2}
\begin{tabular}{@{} l l S[table-format=1.4] S[table-format=1.4] S[table-format=1.4] @{}} 
\toprule
\textbf{Method}  & \textbf{Model}        & {\textbf{R@1}} & {\textbf{R@5}} & {\textbf{R@10}} \\
\midrule
\multirow{2}{*}{VFM} & DINOv2 CLS          & {78.21}          & {\textbf{90.95}}          & {\textbf{94.54}}           \\
                 & DINOv2 GeM          & {77.77}          & {87.94}          & {91.01}           \\
\midrule
\multirow{4}{*}{LLM-VPR} & \multicolumn{4}{l}{\textit{Guidance-Based}} \\
\cmidrule(l{2pt}r{2pt}){2-5}
                 & \quad Qwen-7B       & {81.60}         & {90.29}         & {92.18}          \\
                 & \quad Qwen-7B-UASC  & {\textbf{82.33}}         & {90.92}         & {92.56}          \\

\bottomrule
\end{tabular}
\end{table}

\par
As presented in Table \ref{tab:pitts30k}, the Guidance-based method also delivers notable improvements on the Pitts30k dataset. Qwen-7B surpasses the best-performing VFM baseline (DINOv2 CLS) by 3.39 p.p. in R@1, and Qwen-7B-UASC builds upon this with an additional 0.73 p.p. gain.
\par
We note that the performance gains are concentrated at R@1, with limited improvement for R@5 and R@10. This is an inherent limitation of the two-stage pipeline: the LLM re-ranker cannot recover a correct match if it is absent from the top-20 candidates retrieved by the VFM stage. The performance of our method is therefore bounded by the recall of the initial retrieval.

\subsection{Computational Efficiency}
As detailed in Table \ref{tab:efficiency_comparison_final}, our Guidance-based method achieves remarkable gains in computational efficiency. The reported metrics represent averages over 10 runs, where we measured the processing time and output tokens for each sample and computed the mean values. Compared to the Description-based baseline, our approach accelerates inference by a factor of $\sim$210 for Qwen-7B and $\sim$259 for Qwen-32B. This massive speedup is accompanied by a drastic reduction in generative cost, slashing the number of output tokens by 174-fold and 210-fold for Qwen-7B and Qwen-32B, respectively. This leap in performance and efficiency brings the on-board deployment of multimodal LLMs for VPR tasks on resource-constrained robots closer to a practical reality.

\begin{table}[t]
\centering
\caption{Efficiency Comparison on Tokyo247}
\label{tab:efficiency_comparison_final}
\renewcommand{\arraystretch}{1.2}
\begin{tabular}{@{} l S[table-format=2.1]
                           S[table-format=4.2]
                           S[table-format=5.0, group-separator={,}] @{}} 
\toprule
\textbf{Method} & {\textbf{R@1}} & {\textbf{Avg. Time}} & {\textbf{Avg. Output}} \\
                &                & {\textbf{(s/sample)}} & {\textbf{(Tokens/sample)}} \\
\midrule

\multicolumn{4}{l}{\textit{Description-Based}} \\
\cmidrule(l{2pt}r{2pt}){1-4}
\quad Qwen-7B  & {71.6} & {772.95} & {19499} \\
\quad Qwen-32B & {74.6} & {2576.34} & {27364} \\[0.5ex]

\cmidrule(l{2pt}r{2pt}){1-4}

\multicolumn{4}{l}{\textit{Guidance-Based (ours)}} \\
\cmidrule(l{2pt}r{2pt}){1-4}
\quad Qwen-7B       & {83.8} & {\textbf{3.67}} & {\textbf{112}} \\
\quad Qwen-7B-UASC  & {84.4} & {15.92} & {745} \\
\quad Qwen-32B      & {89.5} & {9.94} & {130} \\
\quad Qwen-32B-UASC & {\textbf{91.1}} & {46.69} &  {826} \\
\bottomrule
\end{tabular}
\end{table}

In our implementation of the UASC strategy, we set the number of sampling iterations to N=5. While this incurs a computational overhead of approximately five-fold, it enables a further enhancement in the R@1 score. Investigating more cost-effective sampling schemes is therefore a valuable direction for future work.


\section{Conclusion}

In this paper, we attempted to address the fundamental conflict between the need for generalization in VPR and the limitations of existing fine-tuning-based or computationally expensive zero-shot methods. We introduced a novel training-free framework, which leverages TTS to guide MLLMs for efficient and robust VPR. Our Guidance-based approach, built on Chain-of-Thought and Self-Consistency, bypasses the information bottleneck of previous description-based methods by enabling direct, end-to-end multimodal reasoning. Our extensive experiments demonstrated the superiority of this approach. On benchmarks like Tokyo247 and Pitts30k, our method not only achieved a remarkable R @ 1 of 91. 11\% in Tokyo247, but also yielded staggering computational efficiency gains of up to 210 times. This leap in efficiency significantly enhances the feasibility of deploying powerful MLLMs on resource-constrained platforms, such as robots, for real-world navigation tasks. Our work represents a significant step towards creating a new class of efficient, adaptable, and high-performance VPR systems built upon the zero-shot capabilities of MLLMs.

\bibliography{ifacconf}             
                                                   







\newpage
\appendix
\begin{figure*}[t!]

\begin{tcolorbox}[
    colback=gray!5,
    colframe=gray!75!black,
    title=\textbf{Appendix A: Complete MLLM Prompt for VPR Re-ranking},
    fonttitle=\bfseries,
    arc=2mm,
    boxrule=1pt,
    breakable 
]

\begin{multicols}{2}


\subsection*{System Instruction}
You are an expert in visual place recognition, functioning as a highly precise and efficient ranking engine. Your core purpose is to compare images of places and determine their similarity based on permanent, static features. You must always ignore transient elements like vehicles, people, animals, weather, and lighting. Your responses must be structured, concise, and directly address the user's query format.

\vspace{0.2cm}
\hrule
\vspace{0.2cm}

\subsection*{User Prompt}
Your task is to meticulously analyze a query image against a candidate image and output a numerical similarity score.

Your internal thought process must follow these steps, but you \textbf{MUST NOT} output this process. This is your internal checklist:

\begin{enumerate}[leftmargin=*, label=\arabic*., topsep=3pt, itemsep=0pt]
    \item \textbf{Strictly Adhere to Constraints:}
    \begin{itemize}[leftmargin=*, label=\textbullet, topsep=2pt, itemsep=0pt]
        \item Your analysis must be based \textbf{only} on permanent, static objects (e.g., buildings, permanent signs, street furniture, architectural features).
        \item You \textbf{must ignore} transient elements.
    \end{itemize}

    \item \textbf{Systematic Object-by-Object Analysis:}
    \begin{itemize}[leftmargin=*, label=\textbullet, topsep=2pt, itemsep=0pt]
        \item Identify all distinct, permanent objects in the query image.
        \item For each object, silently search for a corresponding object in the candidate image.
        \item Internally compare objects based on key attributes: color, shape, structure, patterns, textures, spatial relationships, and text.
        \item Carefully examine details of matched static objects to confirm they are truly the same, accounting for viewpoint changes.
    \end{itemize}

    \item \textbf{Calculate a Similarity Score:}
    \begin{itemize}[leftmargin=*, label=\textbullet, topsep=2pt, itemsep=0pt]
        \item Calculate a score between 0.0 and 1.0 based on matched vs. mismatched permanent objects.
        \item Use the following rubric:
        \begin{description}[font=\normalfont\itshape, topsep=2pt, itemsep=0pt]
            \item[1.0:] Perfect or near-perfect match.
            \item[0.8-0.99:] Very strong match. Definitely the same place.
            \item[0.5-0.79:] Probable match. Several significant landmarks match.
            \item[0.2-0.49:] Weak or partial match. Shares only generic landmarks.
            \item[0.0-0.19:] No significant match. Clearly different locations.
        \end{description}
    \end{itemize}

    \item \textbf{Detailed Inspection for High Scores:}
    \begin{itemize}[leftmargin=*, label=\textbullet, topsep=2pt, itemsep=0pt]
        \item If score > 0.8, perform a more fine-grained comparison.
        \item Pay closer attention to geometric features and texts of matching static objects.
    \end{itemize}
\end{enumerate}

\textbf{Final Output Instruction:} \\
You \textbf{MUST} provide your response \textit{only} as a single, raw JSON object, without any extra text or markdown. The JSON must follow this schema:

\begin{lstlisting}[style=json_style, basicstyle=\ttfamily\footnotesize, numbers=none]
{
  "similarity_score": float,
  "justification": "A brief, one-sentence summary explaining the score.",
  "key_matching_objects": [
    "List of matched static objects."
  ],
  "key_mismatched_objects": [
    "List of mismatched static objects."
  ]
}
\end{lstlisting}


\end{multicols}
\end{tcolorbox}
\end{figure*}

\begin{figure*}[t!]
\vspace{-1.0cm}
\begin{tcolorbox}[
    colback=gray!5,
    colframe=gray!75!black,
    title=\textbf{Appendix B: Example of UASC JSON Output}, 
    fonttitle=\bfseries,
    arc=2mm,
    boxrule=1pt,
    breakable 
]
\vspace{0.2cm}

\begin{lstlisting}[style=json_style, basicstyle=\ttfamily\footnotesize, numbers=none]
{
  "justification": "Both images show a busy urban street with similar storefronts, signage, and general layout, indicating they are likely taken from the same area.",
  "key_matching_objects": [
    "street lamps",
    "signage boards",
    "pedestrian crossing lines"
  ],
  "key_mismatched_objects": [],
  "similarity_score": 0.854504902432036,
  "uncertainty_metrics": {
    "mean_score": 0.8799999999999999,
    "std_dev": 0.050990195135927834,
    "lambda": 0.5,
    "num_valid_samples": 5
  },
  "query_image": "Query.png",
  "candidate_image": "Top3_True.png",
  "inference_time_seconds": 17.8513,
  "sc_details": [
    {
      "sample_index": 0,
      "raw_output": "```json\n{\n  \"similarity_score\": 0.9,\n  \"justification\": \"Both images show a busy urban street with similar storefronts, signage, and general layout, indicating they are likely taken from the same area.\",\n  \"key_matching_objects\": [\"street lamps\", \"signage boards\", \"pedestrian crossing lines\"],\n  \"key_mismatched_objects\": []\n}\n```",
      "status": "Success",
      "parsed_score": 0.9
    },
    {
      "sample_index": 1,
      "raw_output": "{\n  \"similarity_score\": 0.8,\n  \"justification\": \"Both images show a busy urban street with similar storefronts, signage, and general layout, indicating they are likely taken at the same location.\",\n  \"key_matching_objects\": [\"streetlights\", \"signage\", \"buildings\"],\n  \"key_mismatched_objects\": []\n}",
      "status": "Success",
      "parsed_score": 0.8
    },
    {
      "sample_index": 2,
      "raw_output": "{\n  \"similarity_score\": 0.9,\n  \"justification\": \"Both images show a busy urban street with similar storefronts, signage, and general layout, indicating they are likely taken from the same area.\",\n  \"key_matching_objects\": [\"street lamps\", \"signage boards\", \"pedestrian crossing lines\"],\n  \"key_mismatched_objects\": []\n}",
      "status": "Success",
      "parsed_score": 0.9
    },
    {
      "sample_index": 3,
      "raw_output": "```json\n{\n  \"similarity_score\": 0.95,\n  \"justification\": \"Both images show the same urban street scene with recognizable shops and signage, including '3COINS' and 'Claire's', indicating they are likely taken at the same location.\",\n  \"key_matching_objects\": [\n    \"'3COINS' store sign\",\n    \"'Claire's' store sign\",\n    \"Street lamp design\",\n    \"Pedestrian crossing lines\"\n  ],\n  \"key_mismatched_objects\": []\n}\n```",
      "status": "Success",
      "parsed_score": 0.95
    },
    {
      "sample_index": 4,
      "raw_output": "{\n  \"similarity_score\": 0.85,\n  \"justification\": \"Both images show a busy urban street with similar storefronts, signage, and general layout. Key matching objects include the '3COINS' store sign, 'KEN' sign, and various shop facades with consistent branding.\",\n  \"key_matching_objects\": [\"3COINS\", \"KEN\", \"Shop Facades\"],\n  \"key_mismatched_objects\": []\n}",
      "status": "Success",
      "parsed_score": 0.85
    }
  ]
}
\end{lstlisting}

\end{tcolorbox}
\end{figure*}

\end{document}